\documentclass{article} %
\usepackage{iclr2015,times}
\usepackage{hyperref}
\usepackage{url}
\usepackage{helvet}
\usepackage{courier}
\usepackage{latexsym}
\usepackage{amsmath,amsthm,amssymb}
\usepackage{graphicx}
\usepackage{multirow}
\usepackage{url}
\usepackage{verbatim}
\usepackage{caption}
\usepackage{subcaption}

\usepackage{color}

\newcommand{\R}{\mathbb{R}}
\newcommand{\shortcite}{\citet}
\renewcommand{\cite}{\citep} 

\title{Incorporating Both Distributional and Relational Semantics in Word Representations}

\author{
Daniel Fried\thanks{Currently at the University of Cambridge.}\\
Department of Computer Science\\
University of Arizona \\
Tucson, Arizona, USA \\
\texttt{dfried@email.arizona.edu} \\
\And
Kevin Duh\\
Graduate School of Information Science \\
Nara Institute of Science and Technology \\
Ikoma, Nara, JAPAN \\
\texttt{kevinduh@is.naist.jp} \\
}

\iclrfinalcopy %

\begin{document}
\maketitle

\begin{abstract}
\vspace{-5mm}
We investigate the hypothesis that word representations ought to incorporate both distributional and relational semantics. To this end, we employ the Alternating Direction Method of Multipliers (ADMM), which flexibly optimizes a distributional objective on raw text and a relational objective on WordNet. Preliminary results on knowledge base completion, analogy tests, and  parsing show that word representations trained on both objectives can give improvements in some cases.
\vspace{-5mm}
\end{abstract}

\section{Introduction}
\vspace{-3mm}

We are interested in algorithms for learning {\em vector representations} of words. 
Recent work has shown that such representations can capture the semantic and syntactic regularities of words \cite{mikolov13queen} and improve the performance of various Natural Language Processing systems~\cite{turian10word,wang13sequence,socher13parsing,collobert11scratch}.

Although many kinds of representation learning algorithms have been proposed so far, 
they are all essentially based on the same premise of {\em distributional semantics}~\cite{harris54}.
For example, the models of \cite{bengio03neurallm,schwenk07cslm,collobert11scratch,mikolov13distributed,mnih13word} train word representations using the context window around the word.
Intuitively, these algorithms learn to map words with similar context to nearby points in vector space.  

However, distributional semantics is by no means the only theory of word meaning. {\em Relational semantics}, exemplified by WordNet~\cite{miller95wordnet}, defines a graph of relations such as synonymy and hypernymy~\cite{cruse86semantics} between words, reflecting our world knowledge and psychological predispositions. 
For example, a relation like ``dog is-a mammal'' describes a precise hierarchy that complements the distributional similarities observable from corpora.

We believe {\em both} distributional and relational semantics are valuable for word representations, and investigate combining these approaches into a {\em unified} representation learning algorithm based on the Alternating Direction Method of Multipliers (ADMM) \cite{boyd2011distributed}. Its advantages include (a) flexibility in incorporating arbitrary objectives, and (b) relative ease of implementation. We show that ADMM effectively optimizes the joint objective and present preliminary results on several tasks.

\vspace{-3mm}
\section{Distributional and Relational Objectives}
\vspace{-3mm}
\textbf{Distributional Semantics Objective}:  We implement distributional semantics using the Neural Language Model (NLM) of \shortcite{collobert11scratch}. Each word $i$ in the vocabulary is associated with a $d$-dimensional vector $\mathbf{w}_i \in \R^d$, the word's \emph{embedding}. An $n$-length sequence of words $(i_1, i_2, \dots, i_n)$ is represented as a vector $\mathbf{x}$ by concatenating the vector embeddings for each word, $\mathbf{x} = [\mathbf{w}_{i_1};\mathbf{w}_{i_2}\ldots;\mathbf{w}_{i_n}]$. This vector $\mathbf{x}$ is then scored by feeding it through a two-layer neural network with $h$ hidden nodes:
\begin{math}
    \small
    S_{NLM}(\mathbf{x}) = \mathbf{u}^{\top}(f(\mathbf{Ax} + \mathbf{b}))
\end{math},
where $\mathbf{A} \in \R^{h \times (nd)}$, $\mathbf{b} \in \R^h$, $\mathbf{u} \in \R^h$ are network parameters and $f$ is the sigmoid $f(t) = 1/(1 + e^{-t})$ applied element-wise. 
The model is trained using noise contrastive estimation (NCE) \cite{mnih13word}, where training text is corrupted by random replacement of random words to provide an implicit negative training example, $\mathbf{x_c}$. The hinge-loss function, comparing positive and negative training example scores, is: 
\begin{equation}
    \small
    \label{hinge-loss}
    L_{NLM}(\mathbf{x}, \mathbf{x}_c)=\max(0, 1 - S_{NLM}(\mathbf{x}) + S_{NLM}(\mathbf{x}_c))
\end{equation}
The word embeddings, $\mathbf{w}$, and other network parameters are optimized with backpropagation using stochastic gradient descent (SGD) over n-grams in the training corpus.

\textbf{Relational Semantics Objective}: We investigate three different objectives, each modeling relations from WordNet. 
The Graph Distance loss, $L_{GD}$, enforces the idea that words close together in the WordNet graph should have similar embeddings in vector space. First, for a word pair $(i,j)$, we define a pairwise word similarity $WordSim(i,j)$ as the normalized shortest path between the words' synonym sets in the WordNet relational graph~\cite{leacock1998combining}. Then, we encourage the cosine similarity between their embeddings $\mathbf{v}_i$ and $\mathbf{v}_j$ to match that of $WordSim(i,j)$:
\begin{equation}
    \label{eqn:loss_gd}
    \footnotesize
    L_{GD}(i, j) = \left(\frac{\mathbf{v}_i \cdot \mathbf{v}_j}{||\mathbf{v}_i||_2||\mathbf{v}_j||_2}  - \left[a\times WordSim(i, j) + b\right]\right)^2
\end{equation}
where $a$ and $b$ are parameters that scale $WordSim(i,j)$ to be of the same range as the cosine similarity. Training proceeds by SGD: word pairs $(i, j)$ are sampled from the WordNet graph, and both the word embeddings $\mathbf{v}$ and parameters $a$, $b$ are updated by gradient descent on the loss function.

A different approach directly models each WordNet relation as an operation in vector space. These models assign scalar plausibility scores to input tuples $(v_l, R, v_r)$, modeling the plausibility of a relation of type $R$ between words $v_l$ and $v_r$. 
In both of the relational models we consider, each type of relationship (for example, synonymy or hypernymy) has a distinct set of parameters used to represent the relationship as a function in vector space.
The TransE model of \citet{bordes2013translating} represents relations as linear translations: if the relationship $R$ holds for two words $v_l$ and $v_r$, then their embeddings $\mathbf{v}_l, \mathbf{v}_r \in \R^d$ should be close after translating $v_l$ by a relation vector $\mathbf{R} \in \R^d$:
\begin{equation}
    \label{eqn:transe}
    \small
    S_{TransE}(v_l, R, v_r) = -||\mathbf{v}_l + \mathbf{R} - \mathbf{v}_r||_2
\end{equation}
\citet{socher2013reasoning} introduce a Neural Tensor Network (NTN) that models interaction between embeddings using tensors and a non-linearity function. The scoring function for a input tuple is:
\begin{equation}
    \label{eqn:ntn}
    \small
    S_{NTN}(v_l, R, v_r) = \mathbf{U}^{\top}f\left(\mathbf{v}_l^{\top}\mathbf{W}_R\mathbf{v}_{r} + \mathbf{V}_{R} \begin{bmatrix}\mathbf{v}_l \\ \mathbf{v}_r\end{bmatrix} + \mathbf{b}_R\right)
\end{equation}
where $\mathbf{U} \in \R^{h}$, $\mathbf{W}_R \in \R^{d\times d \times h}$, $\mathbf{V}_R \in \R^{h \times 2d}$ and $\mathbf{b}_R \in \R^{k}$ are parameters for relationship $R$. 
As in the NLM, parameters for these relational models are trained using NCE (producing a noisy example for each training example by randomly replacing one of the tuples' entries) and SGD, using the hinge loss as defined in Eq.~\ref{hinge-loss}, with $S_{NLM}$ replaced by the $S_{TransE}$ or $S_{NTN}$ scoring function.

\textbf{Joint Objective Optimization by ADMM}: We now describe an ADMM formulation for joint optimization of the above objectives. Let $\mathbf{w}$ be the set of word embeddings $\{\mathbf{w}_1, \mathbf{w}_2, \ldots \mathbf{w}_{N'}\}$ for the distributional objective, and $\mathbf{v}$ be the set of word embeddings $\{\mathbf{v}_1, \mathbf{v}_2, \ldots \mathbf{v}_{N''}\}$ for the relational objective, where $N'$ and $N''$ are the vocabulary size of the corpus and WordNet, respectively. Let $I$ be the set of $N$ words that occur in both. 
Then we define a set of vectors $\mathbf{y}$ = $\{\mathbf{y}_1, \mathbf{y}_2, \ldots \mathbf{y}_N\}$, which correspond to Lagrange multipliers, to penalize the difference $(\mathbf{w}_i - \mathbf{v}_i)$ between sets of embeddings for each word $i$ in the joint vocabulary $I$, producing a Lagrangian penalty term:
\begin{equation}
    \label{eqn:augmented_lagrangian}
    \footnotesize
L_{P}(\mathbf{w}, \mathbf{v}) = \sum_{i \in I} \left (\mathbf{y}_i^{\top} (\mathbf{w}_i - \mathbf{v}_i) \right ) + \frac{\rho}{2} \left ( \sum_{i \in I} (\mathbf{w}_i - \mathbf{v}_i)^\top (\mathbf{w}_i - \mathbf{v}_i) \right)
    \end{equation}
    In the first term, $\mathbf{y}$ has  same dimensionality as $\mathbf{w}$ and $\mathbf{v}$, so a scalar penalty is maintained for each entry in every embedding vector. This constrains corresponding $\mathbf{w}$ and $\mathbf{v}$ vectors to be close to each other. The second residual penalty term with hyperparameter $\rho$ is added to avoid saddle points; $\rho$ can be viewed as a step-size during the update of $\mathbf{y}$. 

    This augmented Lagrangian term (Eq. \ref{eqn:augmented_lagrangian}) is added to the sum of the loss terms for each objective (Eq. \ref{hinge-loss} and Eq. \ref{eqn:loss_gd}). Let $\theta = (\mathbf{u}, \mathbf{A}, \mathbf{b})$ be the parameters of the distributional objective, and $\phi$ be the parameters of the relational objective. The final loss function we optimize becomes:
\begin{equation}
    \label{eqn:joint_loss}
    \small
    L = L_{NLM}(\mathbf{w}, \theta) + L_{GD}(\mathbf{v}, \phi) + L_{P}(\mathbf{w}, \mathbf{v})
\end{equation}
The ADMM algorithm proceeds by repeating the following three steps until convergence:\\
(1) Perform SGD on $\mathbf{w}$ and $\theta$ to minimize $L_{NLM}+L_{P}$, with all other parameters fixed.\\
(2) Perform SGD on $\mathbf{v}$ and $\phi$ to minimize $L_{GD}+L_{P}$, with all other parameters fixed.\\
(3) For all embeddings $i$ corresponding to words in both the n-gram and relational training sets, update the constraint vector $\mathbf{y}_i := \mathbf{y}_i + \rho(\mathbf{w}_i - \mathbf{v}_i)$. \\
Since $L_{NLM}$ and $L_{GD}$ share no parameters, Steps (1) and (2) can be optimized easily using the single-objective NCE and SGD procedures, with additional regularization term $\rho \left (\mathbf{w}_i - \mathbf{v}_i \right)$. 

\vspace{-3mm}
\section{Preliminary Experiments \& Discussions}
\vspace{-3mm}

\begin{figure}[t]
    \centering
    \includegraphics[width=0.65\linewidth]{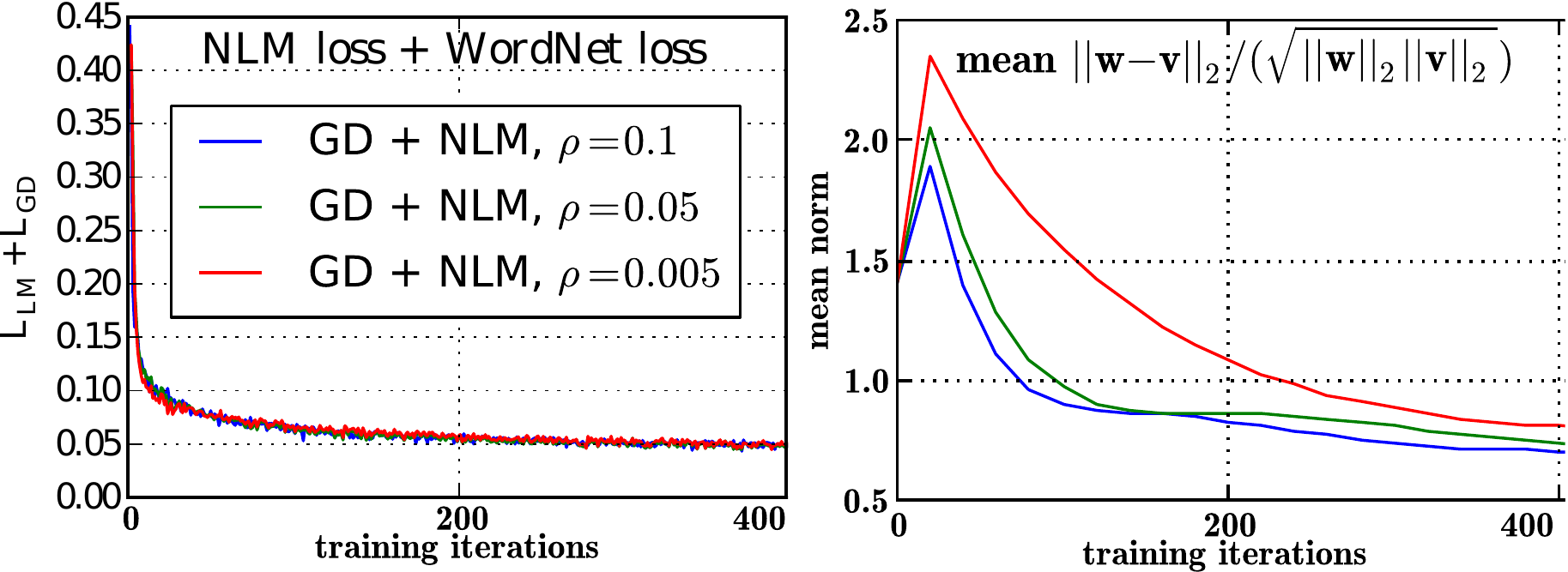}
    \caption{\label{fig:admm_stats} Analysis of ADMM behavior by training iteration, for varying $\rho$. Left: Joint loss, $L_{NLM} + L_{GD}$, on the training data . Right: Normalized residual magnitude, averaged across embeddings. \vspace{-2mm}}
\end{figure}

\begin{table}[t]
    \centering
    \small
    \begin{tabular}{|c|c|cc|cc|cc|}
\hline
                            & NLM & GD & GD+NLM & TransE & TransE+NLM & NTN    & NTN+NLM \\\hline
Knowledge Base & -        &  -    &    -            & 82.87   & {\bf 83.10}      & 80.95  & 81.27 \\ \hline
Analogy Test       & {\bf 42}  & 41 & 41       & 37     & 38               & 36     & 41     \\\hline
Parsing               &  76.03 & 75.90 & {\bf 76.18} & 75.86 & 76.01 & 75.85 & 76.14 \\
\hline
\end{tabular}
\caption{\label{tbl:all} Results summary: Accuracy on knowledge base completion, MaxDiff accuracy on Analogy Test, and Label Arc Score Accuracy on Dependency Parsing for single- and joint-objective models.}
\vspace*{-4mm}
\end{table}

The distributional objective $L_{NLM}$ is trained using 5-grams from the Google Books English corpus\footnote{Berkeley distribution: \url{tomato.banatao.berkeley.edu:8080/berkeleylm_binaries/}}, containing over 180 million 5-gram types. The top 50k unigrams by frequency are used as the vocabulary, and each training iteration samples 100k n-grams from the corpus. For training $L_{GD}$, we sample 100k words from WordNet and compute the similarity of each to 5 other words in each ADMM iteration. For training $L_{TransE}$ and $L_{NTN}$, we use the dataset of~\shortcite{socher2013reasoning}, presenting the entire training set of correct and noise-contrastive corrupted examples one instance at a time in randomized order for each iteration.

We first provide an analysis of the behavior of ADMM on the training set, to confirm that it effectively optimizes the joint objective. 
Fig.~\ref{fig:admm_stats}(left) plots the learning curve by training iteration for various values of the $\rho$ hyperparameter.
We see that ADMM attains a reasonable objective value relatively quickly in 100 iterations. Fig.~\ref{fig:admm_stats}(right) shows the averaged difference between the resulting sets of embeddings $\mathbf{w}$ and $\mathbf{v}$, which decreases as desired.\footnote{The reason for the peak around iteration 50 in Fig.~\ref{fig:admm_stats} is that the embeddings begin with similar random initializations, so initially differences are small; as ADMM starts to see more data, $\mathbf{w}$ and $\mathbf{v}$ diverge, but converge eventually as $\mathbf{y}$ become large.} 

Next, we compare the embeddings learned with different objectives on three standard benchmark tasks (Table \ref{tbl:all}). First, the {\bf Knowledge Base Completion} task \cite{socher2013reasoning} 
evaluates the models' ability to classify relationship triples from WordNet as correct. Triples are scored using the relational scoring functions (Eq.\ref{eqn:transe} and \ref{eqn:ntn}) with the learned model parameters. The model uses a development set of data to determine a plausibility threeshold, and classifies triples with a higher score than the threshold as correct, and those with lower score as incorrect. Secondly, the SemEval2012 {\bf Analogy Test} is a relational word similarity task similar to SAT-style analogy questions \cite{jurgens2012semeval}. Given a set of four or five word pairs, the model selects the pairs that most and least represent a particular relation (defined by a set of example word pairs) by comparing the cosine similarity of the vector difference between words in each pair.
Finally, the {\bf Dependency Parsing} task on the SANCL2012 data~\cite{petrov12sancl} evaluates the accuracy of parsers trained on news domain adapted for web domain. We incorporate the embeddings as additional features in a standard maximum spanning tree dependency parser to see whether embeddings improve generalization of out-of-domain words. The evaluation metric is the labeled attachment score, the accuracy of predicting both correct syntactic attachment and relation label for each word.

For both Knowledge Base and Parsing tasks, we observe that joint objective generally improves over single objectives: e.g. TransE+NLM (83.10\%) $>$ TransE (82.87\%) for Knowledge Base, GD+NLM (76.18\%) $>$ GD (75.90\%) for Parsing. The improvements are not large, but relatively consistent. For the Analogy Test, joint objectives did not improve over the single objective NLM baseline. We provide further analysis as well as extended descriptions of methods and experiments in a longer version of the paper here: \url{http://arxiv.org/abs/1412.4369}.

\section*{Acknowledgments}
This work is supported by a Microsoft Research CORE Grant and JSPS KAKENHI Grant Number 26730121.
D.F. was supported by the Flinn Scholarship during the course of this work. We thank Haixun Wang, Jun'ichi Tsujii, Tim Baldwin, Yuji Matsumoto, and several anonymous reviewers for helpful discussions at various stages of the project. 

\bibliography{mybib}
\bibliographystyle{iclr2015}

\end{document}